# Generation Of Colors using Bidirectional Long Short Term Memory Networks

**Aarush Sinha**

## Abstract

Human vision can distinguish between a vast spectrum of colours, estimated to be between 2 to 7 million discernible shades. However, this impressive range does not inherently imply that all these colours have been precisely named and described within our lexicon. We often associate colours with familiar objects and concepts in our daily lives. This research endeavors to bridge the gap between our visual perception of countless shades and our ability to articulate and name them accurately. A novel model has been developed to achieve this goal, leveraging Bidirectional Long Short-Term Memory (BiLSTM) networks with Active learning. This model operates on a proprietary dataset meticulously curated for this study. The primary objective of this research is to create a versatile tool for categorizing and naming previously unnamed colours or identifying intermediate shades that elude traditional colour terminology. The findings underscore the potential of this innovative approach in revolutionizing our understanding of colour perception and language. Through rigorous experimentation and analysis, this study illuminates a promising avenue for Natural Language Processing (NLP) applications in diverse industries. By facilitating the exploration of the vast colour spectrum the potential applications of NLP are extended beyond conventional boundaries.

## 1. Introduction

In contemporary Language Models (LMs), exemplified by the likes of GPT [4], proficient in linguistic comprehension and adept at reproducing RGB values associated with pre-established colours or objects closely linked to specific hues (e.g., 'Orange,' 'Ruby,' 'Smoke'), there exists a discernible limitation. These models exhibit constraints when associating any word with a distinct colour, thus impeding the discovery of novel chromatic expressions. It is essential to distinguish this challenge from the conventional approach of instructing LMs to generate random RGB values, as the generated outcomes are often devoid of designated nomenclature and are instead considered points along the colour spectrum.

This research explores a novel methodology to ascribe new denominations to unnamed colours. The inquiry extends to soliciting LMs to generate colour representations for arbitrary words, including terms like 'Pina Colada' and 'Semolina,' signifying a drink and a grain, respectively. Conventional LMs are inclined to respond by asserting that such entities lack RGB values, owing to their classification as food items. Paradoxically, human cognition invariably associates these terms with specific colours due to real-world encounters. Consequently, this model delves into instances where unique or intermediate shades find expression, thereby unlocking opportunities for assigning nomenclature to these shades.

Within the scope of this research, we propose adopting a Bidirectional Long Short-Term Memory (BiLSTM) [5,7,12] network with active learning [8,9] to facilitate the visualization of words as distinct colours. The proposed model functions by accepting a given word as input and generating an associated colour as output. Training this model on an extensive dataset of colours, each paired with its corresponding RGB values, thus establishing a profound understanding of the intricate relationship between lexical entities and their chromatic representations.

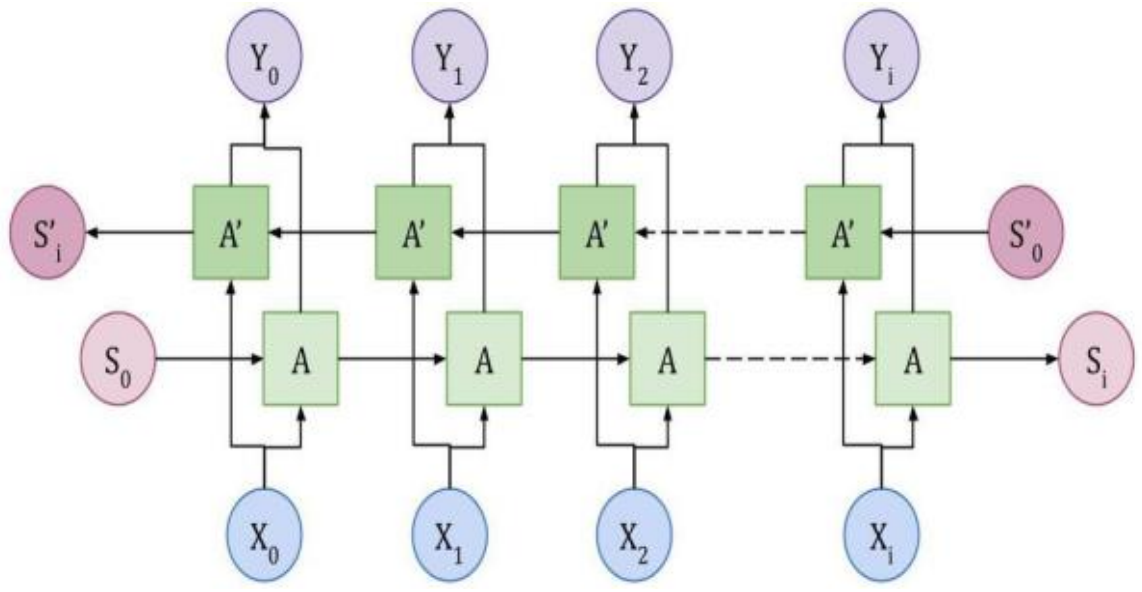

Image 1: Architecture of a BiLSTM

Image 1 describes the architecture of the BiLSTM layer where Xi is the input token, Yi is the output token, and A and A' are LSTM nodes. The final output of Yi is the combination of A and A' LSTM nodes. [6].

## 2. Related Work

Bidirectional Long Short-Term Memory (BiLSTM) [10,11,13] networks have emerged as a prominent and extensively explored architectural innovation in deep learning over recent years. Their remarkable capacity to enhance model performance has been demonstrated across various applications, elevating the standard of natural language processing (NLP) tasks and beyond. This transformative capability is primarily attributed to their unique ability to capture and integrate richer contextual information through bidirectional processing.

The significance of BiLSTMs becomes particularly evident when considering their impact on sequence tagging tasks, such as named entity recognition (NER) [14,15,16] and part-of-speech (POS) tagging [17,18,19]. These tasks demand a nuanced understanding of language, necessitating the identification of not only immediate dependencies but also those that span across a sentence or phrase. BiLSTMs excel in this regard, as they traverse sequences in both forward and backward directions. This dual perspective allows them to effectively encapsulate the intricate relationships between words and their contextual neighbors, resulting in more robust and accurate annotations.

A notable paper by Huang et al 2015 [1] showcases the prowess of BiLSTMs in the context of sequence tagging. Their work introduces the innovative combination of BiLSTMs with Conditional Random Fields (CRFs), a highly effective pairing for structured prediction tasks. By adopting this approach, the model can harness the power of bidirectional information flow to accurately predict labels, achieving state-of-the-art results in tasks like NER. This paper not only underscores the significance of BiLSTMs but also serves as a testament to their ability to transform conventional methodologies.

Similarly, the impact of BiLSTMs extends to neural machine translation (NMT), where they play a pivotal role in elevating translation quality. In NMT, understanding the source language sentence is essential for generating coherent and accurate translations. BiLSTMs, through their bidirectional encoding, empower the model to comprehensively comprehend the source sentence by considering both preceding and succeeding words. This enriched contextual understanding serves as a foundation for producing linguistically accurate and contextually relevant translations.

The seminal work by Bahdanau et al 2014[2] marks a significant milestone in applying BiLSTMs to NMT. While the paper's primary focus revolves around introducing the attention mechanism, it also introduces BiLSTMs as a pivotal architecture component. Using BiLSTMs for source sentence encoding enhances the NMT model's ability to handle complex sentence structures and long-range dependencies. As a result, the quality of translations produced by the model does witness substantial improvements, signifying the transformative potential of BiLSTMs in machine translation.

## 3. Methodology

This section describes the methodology employed to develop and train the Bidirectional Long Short-Term Memory (BiLSTM) network to generate colour representations from text input. The methodology comprises several vital steps: data preprocessing, model architecture definition, training, and prediction. We have also implemented active learning for better outputs and results.

### 4.1 Data Preprocessing

In the first step of our methodology, we prepared the dataset for model training. We set a random seed for reproducibility, loaded the dataset from the specified source, and extracted the colour names from the dataset. To facilitate further processing, we tokenized the colour names and performed padding to ensure uniform length.

### 4.2 Data Normalization

To standardize the RGB values of the colours, we divided the colour data by 255.0, ensuring that all colour channels were in the range [0, 1].

### 4.3 Model Architecture

We constructed a deep-learning model to predict RGB values from colour names. The model consisted of a convolutional layer, a bidirectional LSTM layer, and multiple densely connected layers. The activation functions and layer specifications were carefully chosen to optimize the model's performance. It has 1.3M parameters.

### 4.4 Model Training

With the architecture defined, we compiled the model with an appropriate optimizer and loss function. We then trained the model using the initial data, specifying the number of epochs and batch size. A portion of the data was held out for validation to monitor model performance during training.

Epochs: 350, Batch Size: 512, Validation Split: 20%

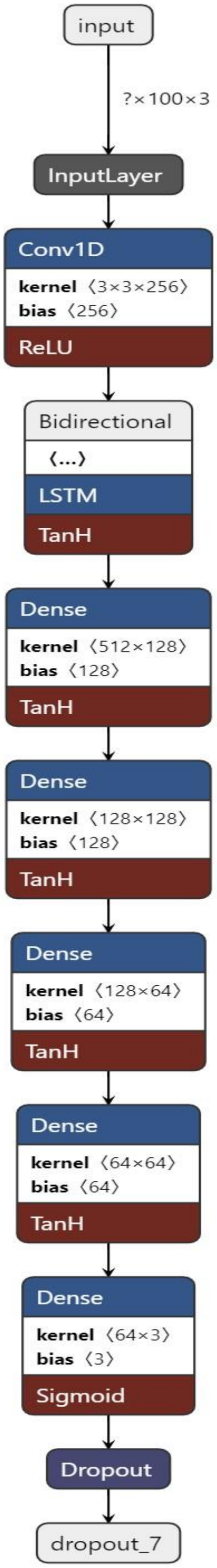

Image 2: Model architecture

4.5 Color Name Generation

To generate new colour names, we combined adjectives and colours. We obtained a list of adjectives from an external source and combined them with a predefined list of colours. This process enabled us to create various colour names for prediction.

4.6 Color Prediction and Dataset Update

A function was implemented to predict RGB values for newly generated colour names. The function utilized the trained model to predict the RGB values and updated the dataset with the new colour name and its corresponding RGB values.

4.7 Active Learning

After adding new data to the dataset, we retrained the model with the updated dataset. This iterative process aimed to improve the model's performance as it gained more data and knowledge about colour names and their corresponding RGB values.

This methodology outlines the critical steps involved in our project, from data preparation to model training, colour name generation, and model refinement. It provides a structured approach to predict RGB values from colour names and continuously enhances the model's accuracy with newly generated data.

Training parameters for second time the model was trained:

Epochs: 15, Batch Size: 1024, Validation Split: 20%

## 5. Results

In the course of our experimental endeavors, the model underwent an exhaustive training regimen harnessing the computational power of a T4 GPU. Preceding the implementation of active learning strategies, the model demonstrated a training accuracy of 73%, accompanied by a validation accuracy of 65%. Subsequent to the integration of active learning techniques, remarkable advancements in model performance were observed: the training accuracy ascended to 75%, while the validation accuracy experienced a substantial upswing, reaching 85%.

The comparatively modest increment in training accuracy is conjectured to be a consequence of the restricted diversity inherent in the input data. Specifically, our dataset is delimited to primary and secondary colors, comprising a mere six distinct hues. As such, the lexicon of adjectives employed for describing these colors inherently offers a finite range of linguistic possibilities. Consequently, we posit that the expansion of our color palette, or a more expansive lexicon encompassing a broader spectrum of descriptors, holds the potential to enhance the model's accuracy and effectively bridge the existing gap between validation and training accuracies post-implementation of active learning.

The substantial surge in validation accuracy stands as a positive testament to the efficacy of active learning. This increase in validation accuracy showcases the model's newfound capability to generalize more effectively to new and unseen data, illustrating the impact of the implemented active learning strategies. This substantiates the potential and efficacy of the techniques employed, thus contributing to the body of knowledge in the field of machine learning and artificial intelligence. It further emphasizes the significance of refining the dataset and lexicon in the pursuit of achieving a more comprehensive and robust model.

Image 3: Graph of training accuracy vs validation accuracy

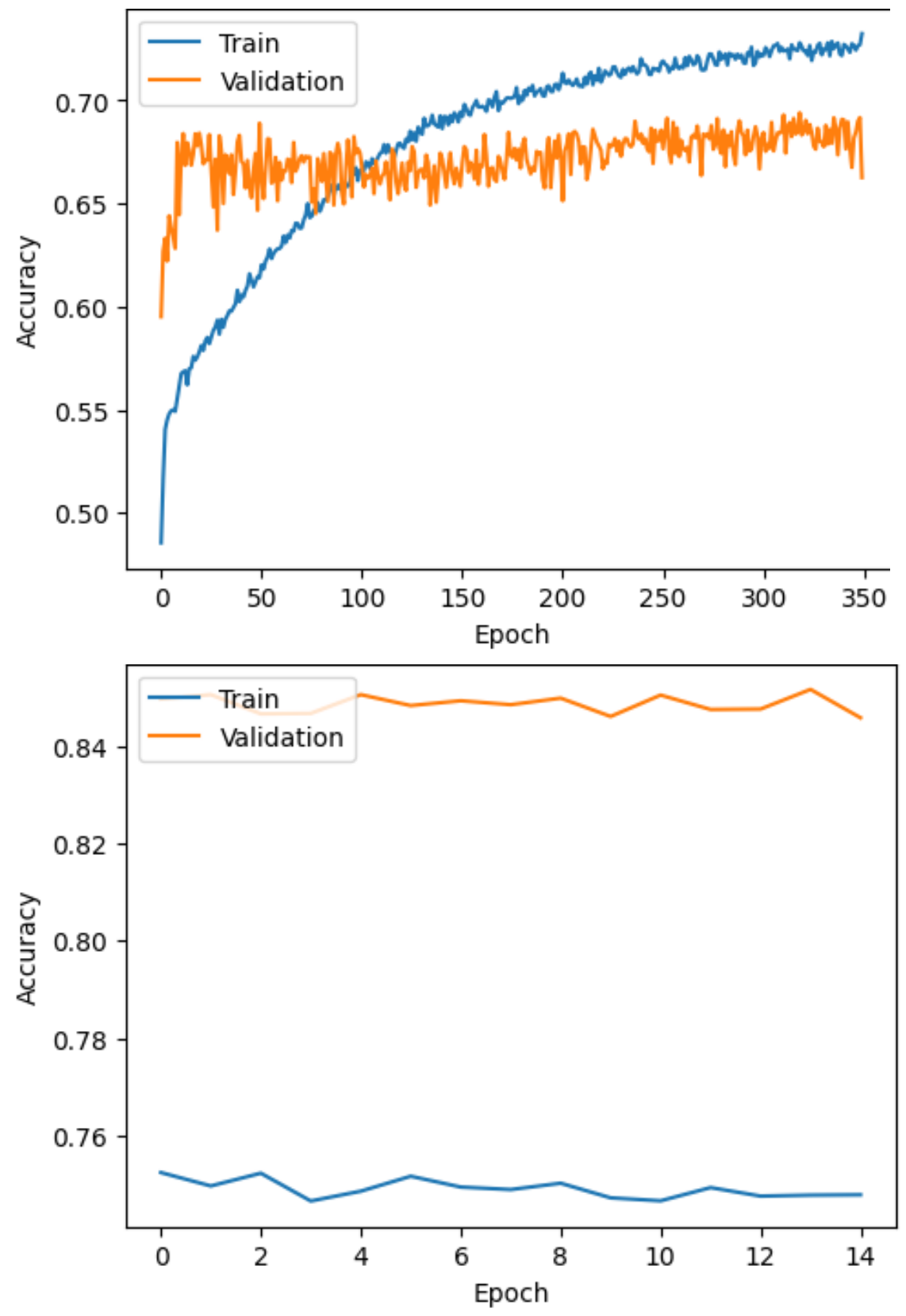

Table 1: Primary and Intermediate Prompts:

| | |
|---|---|
| Red | 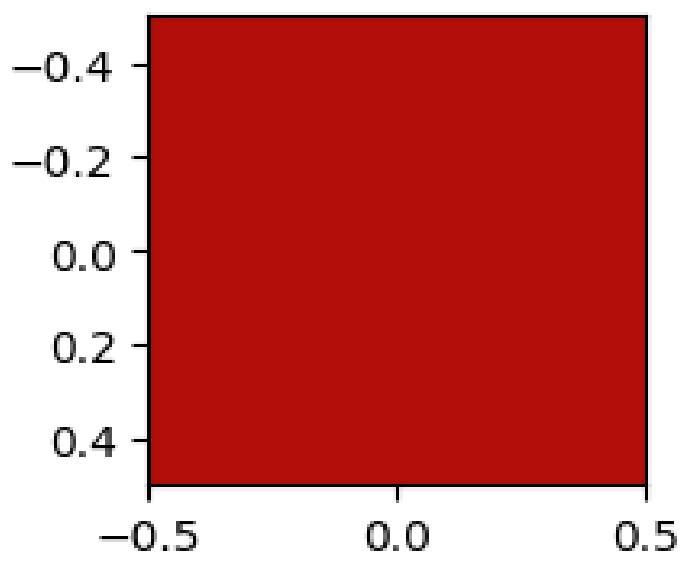  |
| Deep Purple | 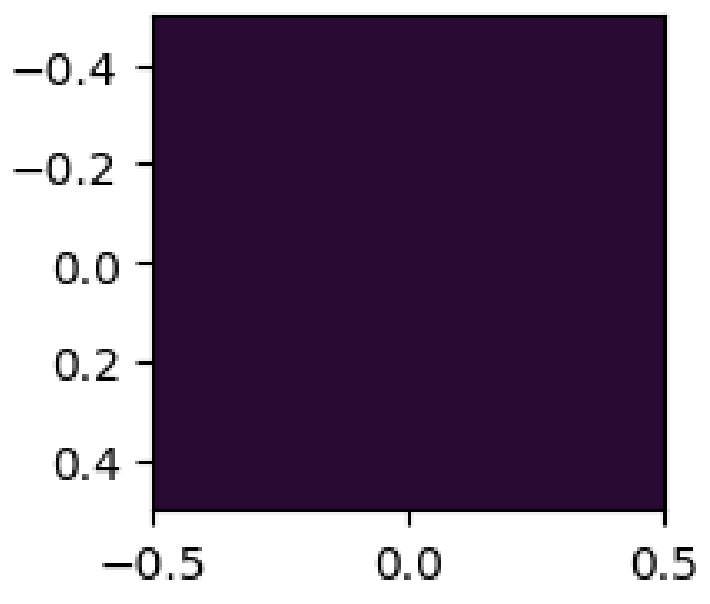  |
| Olive Green | 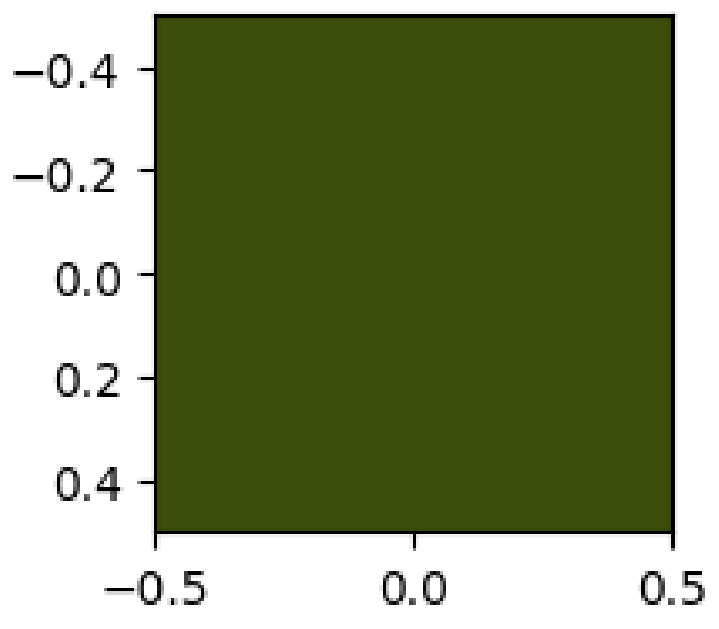  |
| Lemon yellow | 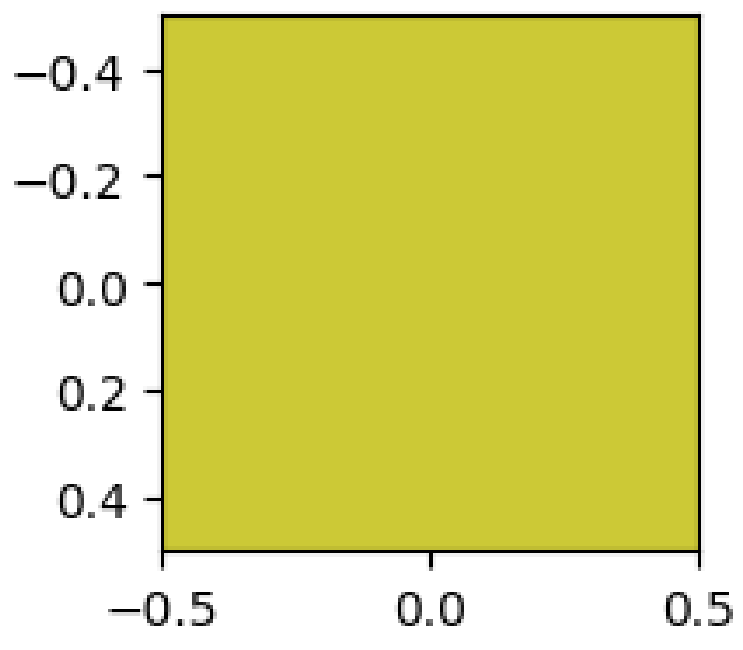  |
| Sky Blue | 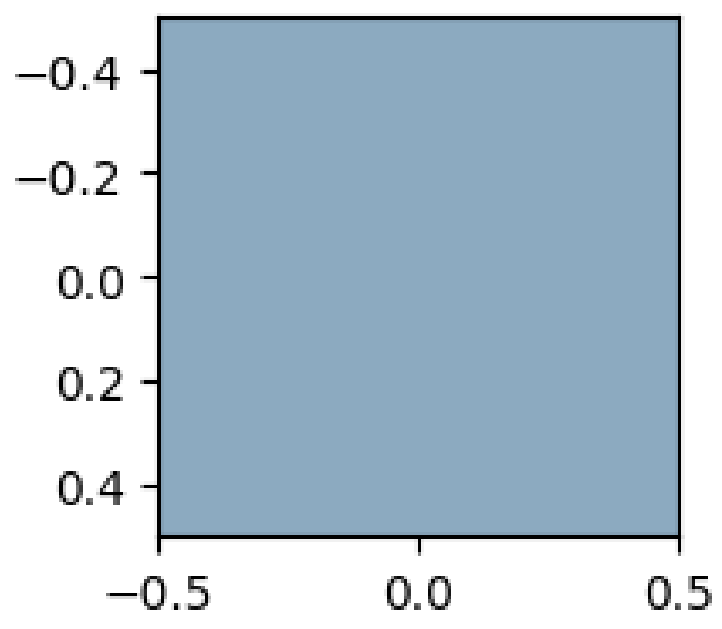 |

Table 2: Abstract Words/Prompts

| | |
|---|---|
| Pina Colada | 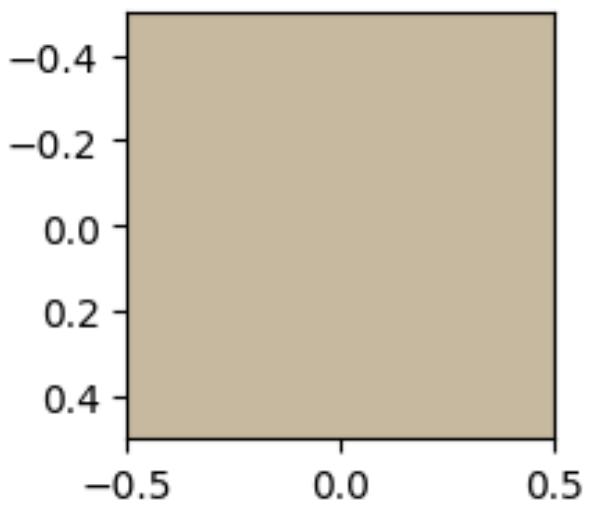 |
| Dragon fruit | 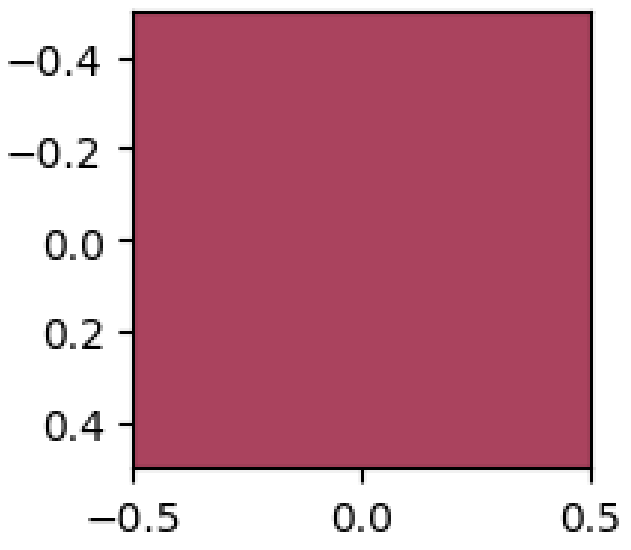 |
| Amethyst | 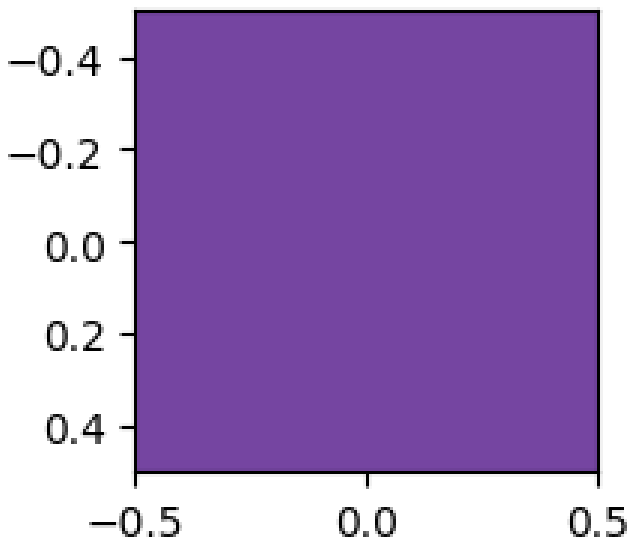 |
| Mohagany | 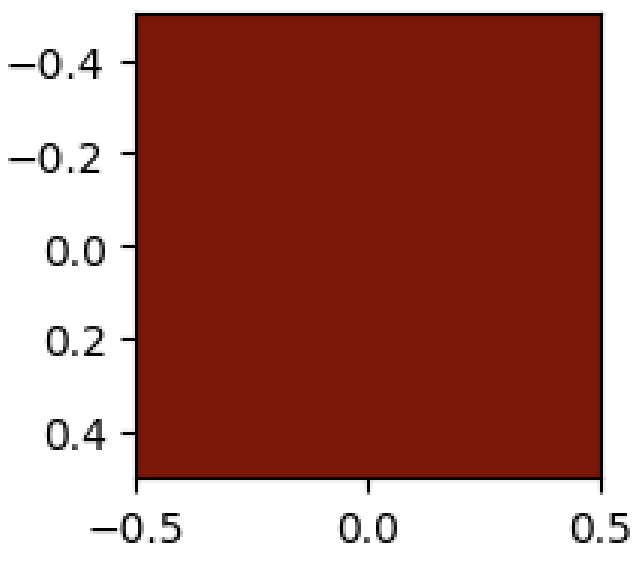 |
| Semolina | 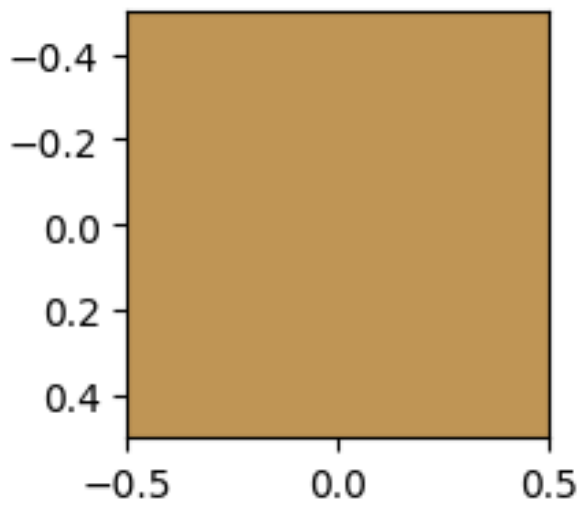 |
| Cucumber | 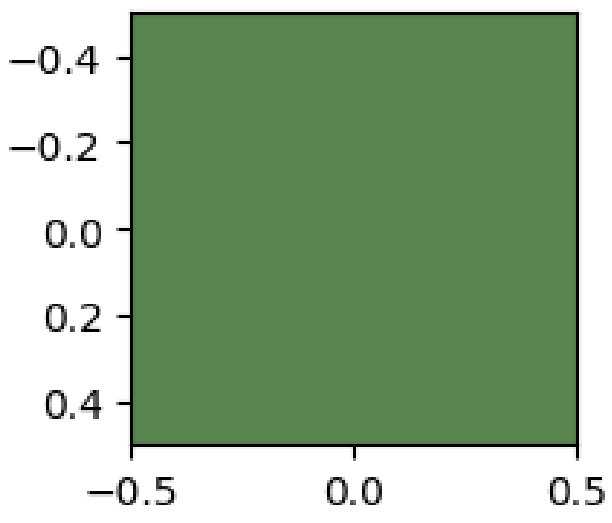 |
| Ocean | 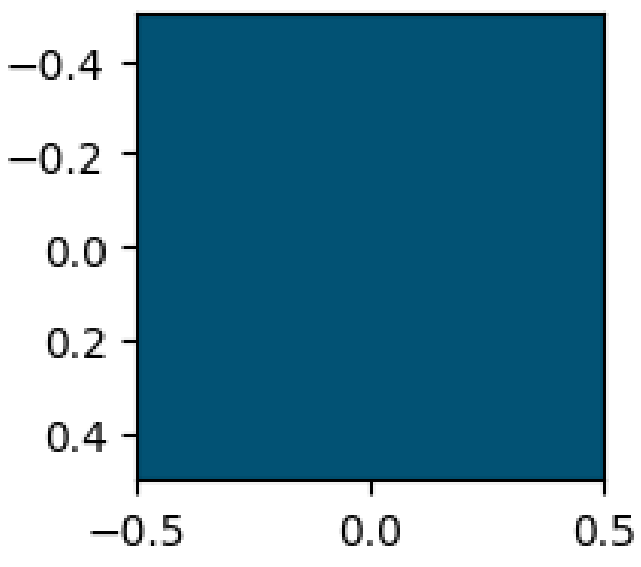 |
| Sand | 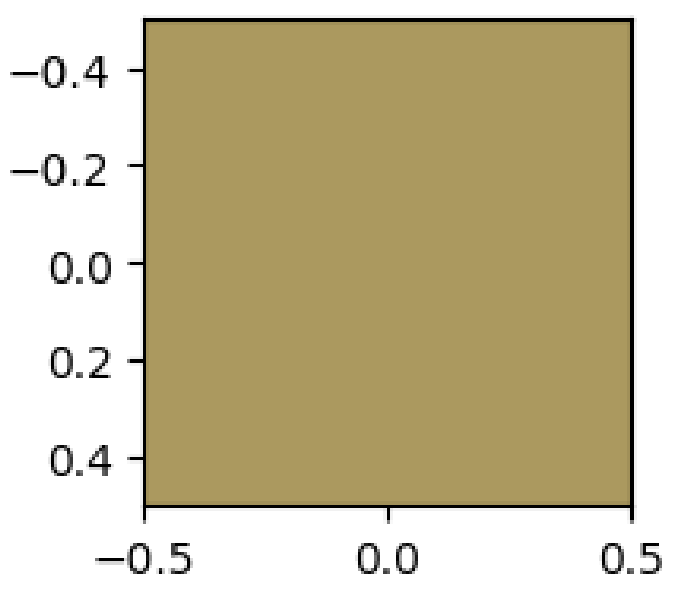 |

Table 3: GPT 3.5 results on prompts given for abstract words

| | |
|---|---|
| RGB Value for Pina Colada | The RGB value for the color "Pina Colada" is not a standardized or universally recognized color, as it's more commonly associated with the name of a tropical cocktail than a specific color. If you have a specific shade or color in mind that you associate with "Pina Colada," you would need to provide additional details or references for a more accurate RGB value. |
| RGB value for semolina | The RGB value for "semolina" is not a standard or predefined color, as it typically refers to a type of wheat flour rather than a specific color. If you have a particular shade or color in mind related to semolina or if you are trying to represent a specific color in a context related to semolina (such as the color of semolina flour), you would need to provide additional information or references for a more accurate RGB value. |
| RGB value for sand | Red: 194 Green: 178 Blue: 128<br><br>Keep in mind that these values are just an approximation, and the appearance of "sand" can differ based on factors like lighting and environmental conditions. To get a precise RGB value for a particular shade of sand, it's recommended to use a color picker tool or consult a color swatch guide. |
| RGB value for cucumber | Red: 144 Green: 200 Blue: 120<br><br>Please note that these values are approximate, and the actual color of a cucumber can vary. If you have a specific shade or context in mind, it's a good idea to use a color picker tool or consult a color reference for a more precise RGB value. |
| RGB value for ocean | The RGB value for the color "ocean" is not a standardized or universally recognized color, as "ocean" typically refers to a body of water and can take on a variety of shades depending on factors such as water depth, location, and lighting conditions. If you have a specific shade or context in mind related to "ocean," you would need to provide additional details or references for a more accurate RGB value. |
| RGB value for mahogany | Red: 192 Green: 64 Blue: 0<br><br>Please note that this is a general approximation, and the RGB value may vary depending on the color palette and definition used |
| | by different sources or applications. For precise color matching, it's advisable to use a color swatch guide or consult specific color references. |
| RGB value for dragon fruit | Red: 226 Green: 58 Blue: 125<br><br>Keep in mind that the actual color of a dragon fruit may vary in different lighting conditions and based on the specific variety, so these values are approximate and can serve as a general reference. |

While the model demonstrates the capability to generate RGB codes for a range of inputs, it should be noted that for specific inputs, such as "Pina Colada", "Ocean", and "Semolina," the model has responded with references to these terms as a drink, a water body with different factors playing in and a grain of wheat, respectively. Moreover, the RGB values provided are presented as approximations, and it is recommended to employ a colour picker tool for precise representation. An additional challenge pertains to colour swatch guides, which, while valuable for colour selection, may not align with the task of text-to-RGB conversion, as they do not accommodate textual input and rely on subjective colour blending preferences.

## 5. Discussions

While our dataset represents a substantial step forward, significant room remains for further enhancement. One promising avenue for improvement lies in the generation of entirely new colour names for RGB values. This endeavor holds the potential to expand the dataset and enhance its richness and diversity. With continued development and fine-tuning, we aspire to achieve a level of accuracy that enables the incorporation of new data directly through the model's outputs.

This forward-looking perspective aligns with our overarching research goals, which extend beyond the present dataset. By pushing the boundaries of colour naming and representation, we aim to refine existing datasets and facilitate the automatic generation of new colour-name pairings, thereby advancing the field of colour recognition and opening doors to novel applications in Natural Language Processing.

## 6. Reproducibility

https://github.com/chungimungi/Color-prediction

## 8. Acknowledgments

The author would like to thank Sravan for their valuable insights on the papers results.